\documentclass[journal]{IEEEtran}
\usepackage{url,hyperref,graphicx,natbib,amsmath,booktabs}
\usepackage[ruled,vlined,linesnumbered]{algorithm2e}

\renewcommand{\cite}[1]{\citep{#1}}

\title{An Incremental Inverse Reinforcement Learning Approach for Motion Planning with Separated Path and Velocity Preferences}
\author{Armin Avaei\,$^{\dagger,1}$, Linda van der Spaa\,$^{\dagger,1,2,*}$, Luka Peternel\,$^{1}$ and Jens Kober\,$^{1}$
\thanks{$^\dagger$These authors contributed equally to this work}
\thanks{$^{1}$Department of Cognitive Robotics, Delft University of Technology, Delft, The Netherlands}
\thanks{$^{2}$Honda Research Institute Europe, Offenbach/Main, Germany}
\thanks{$^*$Corresponding author, email: \texttt{l.f.vanderspaa@tudelft.nl}}}
\renewcommand\footnotemark{}
\date{}

\begin{document}
% \twocolumn
\maketitle

\begin{abstract}
Humans often demonstrate diverse behaviors due to their personal preferences, for instance, related to their individual execution style or personal margin for safety. In this paper, we consider the problem of integrating both path and velocity preferences into trajectory planning for robotic manipulators. We first learn reward functions that represent the user path and velocity preferences from kinesthetic demonstration. We then optimize the trajectory in two steps: first the path and then the velocity, to produce trajectories that adhere to both task requirements and user preferences. We design a set of parameterized features that capture the fundamental preferences in a pick-and-place type of object-transportation task, both in shape and timing of the motion. We demonstrate that our method is capable of generalizing such preferences to new scenarios. We implement our algorithm on a Franka Emika 7-DoF robot arm, and validate the functionality and flexibility of our approach in a user study. The results show that non-expert users are able to teach the robot their preferences with just a few iterations of feedback.
\end{abstract}

%%%%%%%%%%%%%%%%%%%%%%%%%%%%%%%%%%%%%%%%%%%%%%%%%%%%%%%%%%%%%%%%%%%%%%%%%%%%%%%%%%%%%%%%%%%%%%%%%%%%%%%%%%%%%%%%%%%%%%%%%%%%%%%%%%%%%%%%%%%%%%%%%%%%%%%%%%%
\section{Introduction}
Autonomy is increasingly being discussed under the aspect of cooperation. A gentler breed of robots, ``cobots'', have started to appear in factories, workshops and construction sites, working together with humans. A challenge in the deployment of such robots is producing desirable trajectories for object-carrying tasks. A desirable trajectory not only meets the task constraints (e.g.\ collision-free movement from start to goal), but also adheres to user preferences. Such preferences may vary between users, environments and tasks. It is infeasible to manually encode them without exact knowledge of how, with whom, and where the robot is being deployed \cite{jain2015learning}. Manual programming is even more detrimental in cooperative environments, where robots are required to be easily and rapidly reprogrammed. In this context, learning preferences directly from humans emerges as an attractive solution.
\begin{figure}
  %\centering
  \includegraphics[width=\columnwidth]{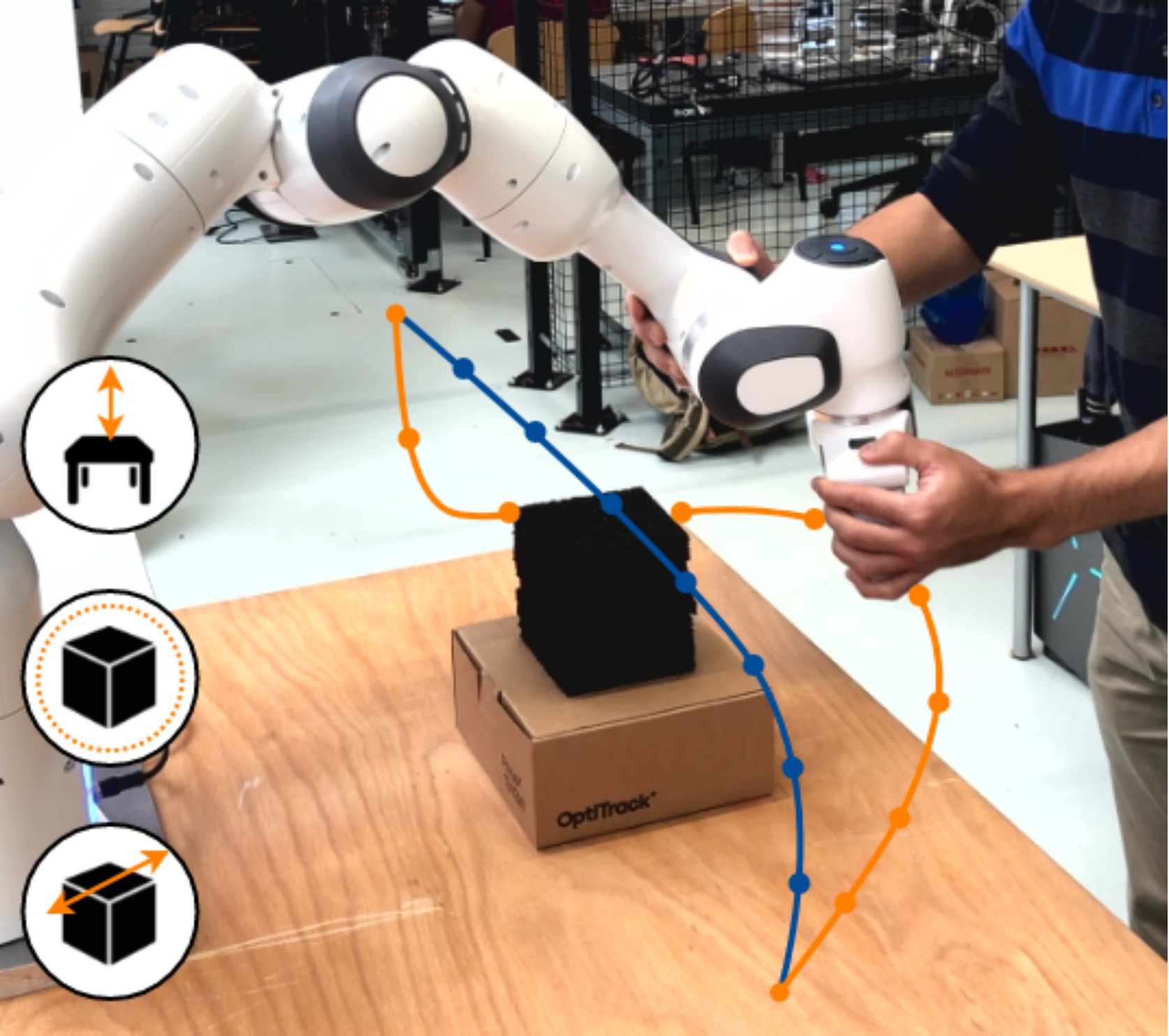}
  \caption{Leveraging demonstrations as means of understanding the human's preferences in an object-carrying task: The robot originally plans the blue trajectory without knowledge of human preferences. The user demonstrates the orange trajectory which in this instance contains the following preferences: ``Stay close to the table surface", ``Keep larger distance from the obstacle", and ``Pass on the far side of the obstacle". We develop a method for learning and generalizing such preferences to new scenarios (i.e.\ new start, goal or obstacle positions).}\label{concept}
\end{figure}

We address the challenge of learning personalized human preferences, starting from a robot plan that may not match the execution style or safety standards of a specific human user (e.g.\ robot carries the object closer to the obstacle than the user prefers). Fig. \ref{concept} illustrates how a user may demonstrate a trajectory encoding multiple implicit preferences to correct the original robot plan. 

One way to adhere to human preferences is by means of variable impedance control \cite{duchaine2007general,peternel2014teaching}. While such strategies can ensure safe and responsive adaptation, they suffer from being purely reactive (i.e.\ they do not remember the corrections). The robot should not only conform to a new trajectory, but it has to update its internal model in order to understand the improvement of the corrected trajectory \cite{bajcsy2017learning,losey2019learning,losey2022physical}. Thus, ideally, we should encode knowledge of humans' desired trajectories as a set of parameters that are incrementally updated based on the corrected trajectory. 

To this end, Learning from Demonstration (LfD) approach enables robots to encode human-demonstrated trajectories. LfD frameworks have the advantage of enabling non-experts to naturally teach trajectories to robots. A widespread trajectory learning method in LfD is Dynamic Movement Primitives (DMPs) \cite{ijspeert2002movement}. In addition to encoding trajectories, DMPs are able to adapt the learned path by updating an interactive term in the model \cite{kulvicius2013interaction,gams2016adaptation}. Additionally, they can adapt the velocity of the motion by estimating the frequency and the phase of a periodic task \cite{peternel2014teaching}, or learning a speed scaling factor \cite{nemec2018human}. As a result, DMPs can capture human path and velocity preferences on a trajectory level. \citet{losey2019learning} demonstrated that such velocity preferences can also be learned online, from interactive feedback, although with some effort. However, these methods lack any knowledge about the task context or why the trajectory was adjusted in the first place. Hence, such an approach fails to generalize user preferences to new scenarios due to the lack of a higher-level understanding of human actions.

A better approach is to pair parameters with features that capture contextual information (e.g.\ distance to obstacle), and utilizes this information to find an optimal solution in new scenarios. Such generalization can be achieved by learning a model of what makes a trajectory desirable. Modeling assumptions can be made to form a conditional probability distribution over trajectories and contextual information, e.g.\ as in \citet{ewerton2016incremental}. While proven effective in simple reaching tasks, whether such models can directly capture complex human preferences in a contextually rich environment remains an open question. However, Inverse Reinforcement Learning (IRL) approaches have already proven to be capable of this \cite{wirth2017survey}.

Unlike traditional IRL methods requiring expert demonstrations \cite{ratliff2006maximum,ziebart2008maximum}, more recently derived algorithms allow preference learning from user comparisons of sub-optimal trajectories \cite{wirth2017survey}. Potentially, a much wider range of human behavior can be interpreted as feedback for preference learning in general \cite{jeon2020reward}. In this paper, however, we focus on reward learning for robot trajectories. A model-free approach can be used to learn complex non-linear reward functions \cite{ibarz2018reward}, but such an approach requires many queries to learn from, which is time-intensive. Therefore, we keep a simple linear reward structure. To shape this reward, we identified four fundamental preference features of the pick-and-place type of object-transportation tasks in the literature: height from table/ground \cite{jain2015learning,bajcsy2017learning,losey2022physical}, distance to obstacle \cite{jain2015learning,biyik2021learning}, obstacle side \cite{kirby2009companion,kretzschmar2016socially}, and velocity \cite{peternel2014teaching,nemec2018human}. These features are relatively scenario-unspecific, and therefore suitable for generalization in object-transportation tasks of the kind we consider in this paper: pick-and-place tasks in the presence of obstacles. To the best of our knowledge, there is no method to account for all these features together in a unified framework.

Given such a set of features, coactive learning \cite{shivaswamy2015coactive} can be used to learn a reward function. In coactive learning, the learner and the teacher both play an active role in the learning process: the learner proposes one or multiple solutions and learns from relative feedback provided by the teacher in response. Coactive learning has an upper boundary on regret, leaving room for noisy and imperfect user feedback. Furthermore, it is an online algorithm, i.e., the system can learn incrementally from sequential feedback. An adapted version of coactive learning was applied in \citet{jain2015learning} to learn trajectory preferences in object-carrying tasks. To this end, users iteratively ranked trajectories proposed by the system. Although selected based on the learned reward, the trajectories were generated using randomized sampling, which increases the number of feedback iterations necessary for convergence. Methods in \citet{bajcsy2017learning,losey2022physical} adapt the robot trajectory to a user's preferences based on force feedback and optimize the remaining trajectory with online correction in a specific scenario. However, these methods cannot capture velocity preferences on top of path preferences.

To address this gap in the state-of-the-art, we propose a novel framework for optimizing trajectories in object-transportation tasks that meet the user's path and velocity preferences, where we first optimize the path and then the velocity on the path. The objective function for the optimization comprises a human preferences reward function and a robot objective function that ensures the safety and efficiency of the trajectories. This explicit separation of the agents' objectives allows for negotiation, where the robot is recognized as an intelligent agent which may give valuable input of its own.

The approach takes a full demonstrated trajectory as the feedback for the learning model, comparing it to the robot's previous plan at each iteration. A minimum acceleration trajectory model significantly reduces the size of the task space, hence increasing the optimization efficiency. 
To capture the preferences, we design a set of features that correspond to the four preferences, covering both the motion shape and timing, which we identified from the literature to be fundamental for the considered pick-and-place tasks.

Unlike \citet{bajcsy2017learning,losey2022physical}, we request iterative feedback and employ an optimization scheme that samples from the global trajectory space. While this is less efficient in terms of human effort for teaching preferences in a specific scenario (i.e.\ the user has to provide at least one full task demonstration), it allows us to additionally capture velocity preferences on top of the path preferences. Furthermore, our method enables the separation of velocity and path preferences both during the learning and in the trajectory optimization stage. With our combination of a trajectory optimization scheme and carefully selected preference features, we can generalize to new contexts without needing (many) additional corrective demonstrations. In contrast to \citet{jain2015learning}, we learn from a few informative feedback demonstrations, and give special attention to the trajectory sampling by employing model-based trajectory optimization. This facilitates fast learning and generalization of preferences to entirely new contexts. 

We evaluate the proposed method in a user study on a 7-DoF Franka Emika robot arm. In the key previous user studies of learning human preferences \cite{palan2019learning,bajcsy2017learning,losey2022physical}, the experimenter instructed the human participants what preference to demonstrate to the robot. Differently, in our user study, we let the participants freely select their own preferences while demonstrating the task execution to the robot. Additionally, our study examines whether the users can actually distinguish the learned trajectory capturing their preference from the trajectories capturing only part of their preference. 
In a supplementary study, we qualitatively compare our method to two relevant methods from the literature. We discuss the structural differences between the methods, and show in simulation how these differences affect the learning of preferences from human (corrective) demonstrations.

In summary, this paper's main contribution is a methodology that is able to capture velocity preferences on top of path preferences by separating the velocity optimization from the path optimization. Learning the path and velocity separately provides users with the option to avoid the challenge of providing a temporally consistent demonstration at each iteration. This offers users the flexibility to demonstrate their path and velocity preferences either simultaneously or in separate demonstrations. Secondly, the learned preferences are transferred to new scenarios by exploiting a trajectory model. Importantly, we perform a user study to validate whether the proposed method can learn and generalize freely chosen preferences, in contrast to the many user studies in the literature which prescribe user preferences. Additionally, we perform a supplementary study to compare pros and cons of the proposed approach to two common methods from the literature.

The rest of the paper is organized as follows: In Sec.~\ref{methods}, we explain the algorithm and methodology in detail. The user study is described in Sec.~\ref{UserStudy}, and the experimental results are shown and discussed. A supplementary study is presented and discussed in Sec.~\ref{ComparisonStudy}. Finally, we present our conclusion and view on future work in Sec.~\ref{Conclusion}.

%%%%%%%%%%%%%%%%%%%%%%%%%%%%%%%%%%%%%%%%%%
\section{Method}\label{methods}
The problem is defined in the following manner: given a context $\mathcal{C}$ describing start, goal, and obstacle positions, the robot has to determine the trajectory $\boldsymbol{\xi}=[\pmb{s}_1, \pmb{s}_2, \ldots, \pmb{s}_N] \in \Xi$ (set of state sequences) that conforms to the human preferences and meets the task goals. The states are defined as $\pmb{s}_k=[\pmb{x}_k;\dot{\pmb{x}}_k]$ (position and velocity), with $k$ indicating trajectory samples.

In our setting, the true reward functions are known by the user but not directly observable by the robot. Hence the problem can be seen as a Partially Observable Markov Decision Process (POMDP) \cite{bajcsy2017learning}. Our reward functions have parameters that are part of the hidden state, and the trajectories provided by the user are observations about these parameters. Solving such problems, where the control space is very complex and high-dimensional, is challenging. Therefore, we simplify the problem through approximation of the policy by separating planning and control, and treating it as an optimization problem. Furthermore, we make the problem tractable by reducing our state space to one of viable smooth trajectories. 

\begin{figure} 
  %\centering
  \includegraphics[width=\linewidth]{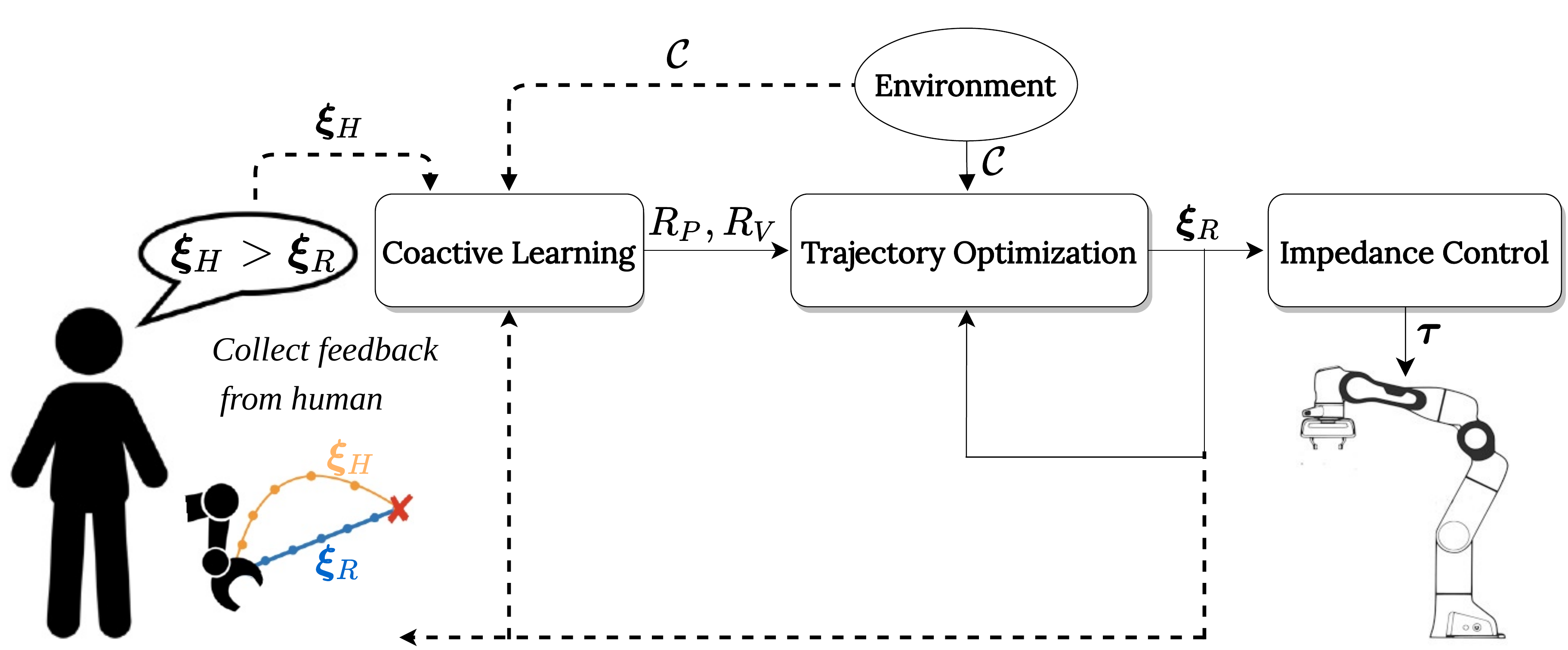}
  \caption{The human user provides demonstrations, which are used to learn a distribution over reward functions via coactive learning. We use the learned rewards to optimize the robot's trajectory according to the human preferences. The resulting trajectory is executed using an impedance controller. We repeat this process, querying the human for preferred trajectories until convergence. The human can then be taken out of the loop.}
  \label{Overview}
\end{figure}

The resulting framework, depicted in Fig.~\ref{Overview}, first learns the appropriate reward functions, then plans a trajectory maximizing the rewards via optimization. Once the trajectory is defined, we use impedance control to track it in a safe manner. Notably, we separate the problem of path and velocity planning in the learning and optimization steps. Updating the path and velocity weights separately provides users with the option to avoid the challenge of providing a temporally consistent demonstration at each iteration. As a result, users have the flexibility to demonstrate their path and velocity preferences either simultaneously or in separate demonstrations.

\subsection{Learning human reward functions from demonstration}\label{LfD}
We follow previous IRL work \cite{jain2015learning,ratliff2006maximum} in assuming that the reward functions are a linear combination of features $\phi$ with weights $\theta$. Accordingly, we define path and velocity reward functions $R_{P}$ and $R_{V}$ as
\begin{subequations}\label{humanRewards}
\begin{align}
    R_{P}(\pmb{x};\mathcal{C},\boldsymbol\theta_{HP})&=\boldsymbol\theta_{HP}^T \boldsymbol\Phi_{P}(\pmb{x};\mathcal{C}),
    \label{reward:path}
    \\
    R_{V}(\bar{\dot{x}},\bar{\pmb{x}};\mathcal{C},\boldsymbol\theta_{HV})&=\boldsymbol\theta_{HV}^T \boldsymbol\Phi_{V}(\bar{\dot{x}},\bar{\pmb{x}};\mathcal{C}),
    \label{reward:velocity}
\end{align}
\label{reward}%
\end{subequations}
where $\boldsymbol\theta_{HP}$ and $\boldsymbol\theta_{HV}$ denote the unknown weights that respectively capture the human path and velocity preferences. In case of the velocity reward, we divide the trajectory into equal segments (i.e.\ range of samples) indicated by $r$. Then, $\bar{\pmb{x}}_r$ and $\bar{\dot{x}}_r$ are the average of the position vectors and the velocity norms in a segment. $\boldsymbol\Phi_{P}$ and $\boldsymbol\Phi_{V}$ are the total path and velocity feature counts along the trajectory:
\begin{equation}\label{PhiPV}
    \boldsymbol\Phi_{P}(\pmb{x};\mathcal{C})=\sum_{k=1}^{N} \boldsymbol\phi_{P} \left(\pmb{x}_k; \mathcal{C}\right),
    ~
    \boldsymbol\Phi_{V}(\bar{\dot{x}},\bar{\pmb{x}};\mathcal{C})=\sum_{r=1}^{M} \boldsymbol\phi_{V} \left(\bar{\dot{x}}_r,\bar{\pmb{x}}_r; \mathcal{C}\right).
\end{equation}
Note that the velocity features are a function of both the segment's velocity and position, allowing us to capture position-dependent velocity preferences. 

To have comparable rewards, all trajectories are re-sampled to contain a fixed number of $N$ states. The velocity inherently affects the number of samples within a trajectory, which is why we divide the trajectory into $M$ segments and consider the average velocity within each segment ($M\!<\!N$). Features are directly computed from the robot state and context of the task. We describe them in the next subsection. 

During kinesthetic demonstration, the robot is in gravity compensation mode. That gives the human full control over the demonstrated trajectories, which we assume to correlate exponentially to the human's internal reward:
\begin{equation}\label{PxiH-proportional}
    P(\boldsymbol\xi_H|\mathcal{C},\boldsymbol\theta_{HP},\boldsymbol\theta_{HV}) \propto e^{\boldsymbol\theta_{HP}^T\boldsymbol\Phi_P(\boldsymbol\xi_H;\mathcal{C})+\boldsymbol\theta_{HV}^T\boldsymbol\Phi_V(\boldsymbol\xi_H;\mathcal{C})},
\end{equation}
which, for brevity, we can write as $P(\boldsymbol\xi_H|\mathcal{C},\boldsymbol\theta_H) \propto e^{\boldsymbol\theta_H^T\boldsymbol\Phi(\boldsymbol\xi_H;\mathcal{C})}$.

Assuming that the human behavior is approximately optimal with respect to the true reward (i.e.\ their preferences), we use a variant of coactive learning introduced in \citet{bajcsy2017learning} to learn the weights $\boldsymbol\theta_{HP},\boldsymbol\theta_{HV}$. However, we can only compute our $\boldsymbol\Phi_{P},\boldsymbol\Phi_{V}$ \eqref{PhiPV} over full trajectories. Therefore, instead of updating the weights based on an estimate of the human's intended trajectory from physical interaction, we use a full kinesthetic trajectory demonstration by the human after each task execution to update the sum of the features over the trajectory \eqref{PhiPV}.
This results in the following incremental update rule:
\begin{equation} \label{Update}
{\boldsymbol\theta}_{H}^{i+1}={\boldsymbol\theta}_{H}^{i}+\alpha\left(\boldsymbol\Phi\left(\boldsymbol\xi_{H}^{i};\mathcal{C}\right)-\boldsymbol\Phi\left(\boldsymbol\xi_{R}^{i};\mathcal{C}\right)\right),
\end{equation}
at iteration $i$, with learning rate $\alpha \in (0,1]$. Intuitively, the update rule is a gradient that shifts the weights in the direction of the human’s observed feature count. It should be noted that we update the path preferences only using the position part of the state, and the velocity preferences are updated depending on where in space the velocities were observed.

\subsection{Features and rewards}
We define the objective function for trajectory optimization as a combination of human rewards and robot objectives. The human rewards consist of features that capture human preferences \eqref{humanRewards}, whereas the robot objectives define a basic behavior for the robot. Moreover, the robot objectives counterbalance the effect of the human rewards in the optimization. While we learn the weights in the human rewards (Sec.~\ref{LfD}). The weights in the robot objectives are hand-tuned. In this section, we first describe the features associated with the human rewards, and then the robot objectives.

The human preferences are captured via the four features listed below (see Fig.~\ref{concept} for an example of the listed path preferences). We chose these features as they characterize dominant behaviors in manipulation applications that depend on user preferences. Additionally, the features cover the different dimensions of the workspace (in space and time), creating a complete definition of motion behavior. 

\textit{\underline{Height from the Table}}: The preferred height from the table, on a range of `low' to `high' is captured by the sigmoid function $\phi_{h}=\frac{1}{1+e^{-\lambda (h + p)}}$ with $h$ indicating the vertical distance from the table, $p$ the center of the function (an arbitrary `medium' height above the table), and $\lambda$ the parameter defining the shape of the function. The choice of a sigmoid function is to hinder the effect of this preference when close to upper and lower boundaries during the weight update (e.g., a demonstration at 75~cm above the table should not impact the weight update very differently from a demonstration at 70~cm). The decreasing slope at the boundaries additionally allows other objectives to have a higher impact on the trajectory in such regions during the optimization.

\textit{\underline{Distance to the Obstacle}}: We encode the user's preferred distance to the obstacle, on a range of `close' to `far' using the exponential feature $\phi_{d} = e^{-\beta d^{2}}$, where $d$ is the Euclidean distance to the center of the obstacle, and $\beta$ is the shape parameter. This exponential function gradually drops to 0 at a certain distance from the obstacle. This distance is a threshold outside which the local behavior of the optimization is no longer affected by the distance to the obstacle. Importantly, if a negative weight is learned associated with this feature, the trajectory is still attracted towards the obstacle even if the initial trajectory lies outside of this threshold. This is because our optimization strategy globally explores different regions of the workspace, and in this case it would detect that there is a reward associated with being closer to the obstacle.

\textit{\underline{Obstacle Side}}: 
We define this feature on a range of `close' (the side of the obstacle closer to the robot) to `far' (the side of the obstacle far from the robot) via the tangent hyperbolic function $\phi_{s}=\frac{2}{1+e^{\gamma S}}-1$. Here, $S$ is the lateral distance between a trajectory sample and the vertical plane at the center of the obstacle, and $\gamma$ is a shape parameter. This symmetric function is designed to have a large span in order to be active in all regions of the workspace. However, as the gradient of this function decreases at larger lateral distances, so does
the influence of this function in the local trajectory optimization.

\textit{\underline{Velocity}}: To encapsulate the user's velocity preferences, we adopt a different approach using a discretized linear combination of uniformly distributed Radial Basis Functions (RBFs) in a range $[\dot{x}_{\min},\dot{x}_{\max}]$. For each segment $r$, we map the average velocity norm $\bar{\dot{x}}_r$ onto these RBFs, given by:
\begin{equation}\label{equ:RBF}
   \psi_{j}\left(\bar{\dot{x}}_r\right) = e^{-\left(\varepsilon \bar{\dot{x}}_r - c_{j}\right)^{2}},
\end{equation}
where shape variable $\varepsilon$ defines the width, and $c_{j}$ defines the center of the $j$\textsuperscript{th} RBF, with $j=1,2,\ldots,n$ (we use $n$=9). 

Inspired by \citet{fahad2018learning}, we discretize the above feature to two bins, based on the distance $d_r$ of each segment center to the obstacle. Hence, we have two cumulative feature vectors: $\boldsymbol\Phi_{V1}$ for $d_r \in [0,d_{c})$, and $\boldsymbol\Phi_{V2}$ for $d_{r} \in [d_{c},\infty)$. This allows us to approximate the speed of motion separately in areas considered to be respectively `close' to or `far' from the obstacle based on the distance threshold $d_{c}$ (obtained from demonstration data). This way, we capture velocity preferences relative to the obstacle position. Similarly, features can be defined relative to other context parameters, to capture velocity preferences that depend on other parameterized positions. 

However, the issue might arise that the two trajectories do not have the same number of segments in each distance bin. In such a case, we employ feature imputation using the mean of the available values.

The robot's objectives are composed of the following:

\textit{\underline{Path Efficiency Reward}}: We calculate the total length of a trajectory, which we use as a negative reward. Penalizing the trajectory length is essential in counterbalancing the human preference features in the optimization process. Essentially, it pulls the trajectories towards the straight line path from start to goal and rewards keeping them short.

\textit{\underline{Collision Avoidance Reward}}: We use the obstacle cost as formulated by \citet{zucker2013chomp}, which increases exponentially once the distance to the obstacle drops below a threshold. The negative cost is our reward.

\textit{\underline{Robot Velocity Reward}}: This reward achieves a low and safe velocity in absence of human velocity preferences and is defined based on \eqref{equ:RBF}. In IRL, it is beneficial to learn how people balance other features against a default reward \cite{vasquez2014inverse}.

\subsection{Motion planning via trajectory optimization}
We discuss the problem of motion planning in two parts. First, we address the optimization of the path of the trajectory in the workspace. We then address the optimization of the velocity along this path, defining the timing of the motion.

Solving the path optimization problem over the Cartesian task-space would be complex and inefficient. Instead, we employ a trajectory planning algorithm \cite{wypt} that interpolates between waypoints with piecewise clothoid curves. This algorithm minimizes the acceleration which results in a smooth and realistic motion. We exploit this algorithm to significantly reduce the search space for the path optimization, and sample trajectories using a vector of waypoint coordinates $\mathbf{p}$ and its corresponding time vector $\mathbf{t}_P$, $\boldsymbol\xi=f(\mathbf{p},\mathbf{t}_{P})$.

We consider three waypoints $\mathbf{p}=[\mathbf{p}^s; \mathbf{p}^m; \mathbf{p}^g]$, corresponding respectively to the start position, an arbitrary position within the path, and the goal position. We further simplify the problem by fixing the time vector to ${\mathbf{t}_{P}=t^{g}[0; \frac{D(\mathbf{p}^m)}{D(\mathbf{p}^{g})}; 1]^T}$, where $D(\cdot)$ indicates the Euclidean distance of a waypoint to $\mathbf{p}^{s}$, and $t^{g}$ is the time, just for the path optimization, we assume all trajectories take to finish\footnote{The shape of the paths is not affected by $t^{g}$ in the time ranges of our manipulations, therefore we assume the path to be independent of velocity.}. An uneven distribution of waypoints would bias the reward value. Setting up the time vector in this manner ensures a constant velocity throughout the trajectory, which results in an even distribution of samples over the path. Trajectories can then be sampled only as a function of waypoint positions $\boldsymbol\xi=f(\mathbf{p})$. 

\begin{figure}
  %\centering
  \includegraphics[width=\columnwidth]{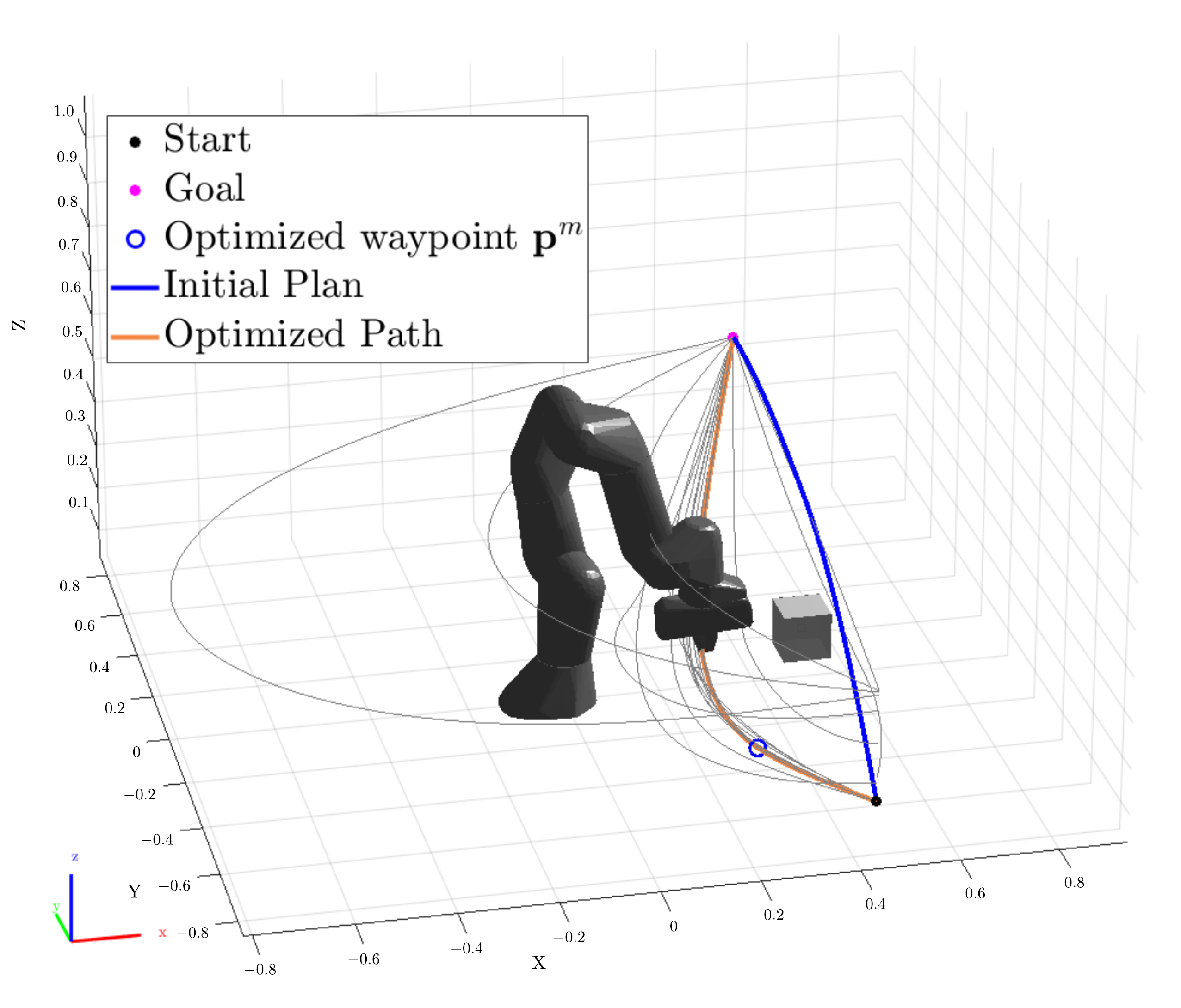}
  \caption{An example of convergence towards the optimal path. The optimizer places $\mathbf{p}^{m}$ in different locations in the workspace to generate different paths. The paths explored by the optimizer are indicated in gray. The orange path indicates the output of the path optimizer, resulting from placing the middle waypoint at the location indicated by the blue circle.}
  \label{convergence}
\end{figure}

We then solve for the optimal waypoint vector $\mathbf{p}^{*}$ using the following non-linear program formulation:
\begin{equation}
\begin{aligned}
&\mathbf{p}^{*}=\arg\max_{\mathbf{p}} \bigg(R_{P}(\mathbf{p};\mathcal{C},\boldsymbol\theta_{HP})+\boldsymbol\theta_{RP}^T \boldsymbol\Phi_{RP}(\mathbf{p};\mathcal{C})\bigg),\\  
&\text {subject to: } \boldsymbol{h}(\mathbf{p})=\mathbf{0}, \ \mathbf{p}_{\mathrm{low}} \leq \mathbf{p} \leq \mathbf{p}_{\mathrm{upp}}.
\end{aligned}
\label{equ:path_optim}
\end{equation}
Here, the objective function consists of the human path reward $R_{P}$ and the robot's path objective, which is a linear combination of predetermined weights $\boldsymbol\theta_{RP}$ and the aforementioned path reward functions $\boldsymbol\Phi_{RP}$. The equality constraint ensures the start and goal positions are met. As a result, we are effectively searching for the waypoint $\mathbf{p}^{m}$ that maximizes the objective function. The upper and lower boundaries $\mathbf{p}_{\mathrm{low}}$ and $\mathbf{p}_{\mathrm{upp}}$ limit the trajectory to stay within the robot's workspace. Once $\mathbf{p}^{*}$ is found, we construct the full trajectory using $\boldsymbol\xi^*_{P}=f(\mathbf{p}^*,\mathbf{t}_{P})$. Fig.~\ref{convergence} shows an example of the convergence of the optimizer towards a path that adheres to `low height', `close side' and `close to obstacle' preferences.

Having the optimal path $\boldsymbol\xi^*_{P}$, we divide the trajectory into $M$ segments (as described in Sec.~\ref{LfD}). Next, we store the positions of the waypoints at the end of the segments in $\mathbf{p}^{*}_{V}=[\mathbf{p}_1, \mathbf{p}_2, \ldots, \mathbf{p}_M]$. This vector is fixed to maintain the shape of the trajectory. The corresponding timestamps, stored in $\mathbf{t}=[t_1, t_2, \ldots, t_M]$, are the variables we optimize. Thus, trajectories sampled by the optimizer are only a function of the time vector $\boldsymbol\xi=f(\mathbf{t})$. By optimizing $\mathbf{t}$ we optimize the average velocity of each segment. The optimal time vector
\begin{equation}
\begin{aligned}
&\mathbf{t}^{*}=\arg\max_{\mathbf{t}} \bigg(R_{V}(\mathbf{t};\mathcal{C},\boldsymbol\theta_{HV})+\theta_{RV} \phi_{RV}(\mathbf{t};\mathcal{C})\bigg),\\  
&\text {subject to: } \boldsymbol{g}(\mathbf{t}) \leq \mathbf{0}, \ \mathbf{t} \leq \mathbf{t}_{\mathrm{upp}},
\end{aligned}
\end{equation}
where the objective function is composed of $R_{V}$ and the robot's velocity objective $\phi_{RV}$, which provides a reward for carrying objects at $\dot{x}_\mathrm{robot}$ with a fixed weight $\theta_{RV}$. The inequality constraint $\boldsymbol{g}(\mathbf{t})$ bounds the velocity over each segment to $\dot{x}_{\min}$ and $\dot{x}_{\max}$, not allowing the timestamps to get too close or far from each other. The upper boundary on $\mathbf{t}$ acts as a limit on the total duration of motion. 

\begin{algorithm}[t!]
Record $\boldsymbol\xi_{H}^{0}=\{\pmb{x}_k,t_{k}\}_{k=1}^N$, obtain context $\mathcal{C}$

$\dot{\pmb{x}}_k \leftarrow \frac{d}{dt}\pmb{x}_k$, compute $\bar{\dot{x}}_r$ and $\bar{\pmb{x}}_r$

Initialize $\boldsymbol\theta^{0}_{H}, \boldsymbol\theta_{R}, \boldsymbol\xi_{R}^{0}$

Set $i=0$

\While{executing task}{ 

\If {Received Human Feedback}{
${\boldsymbol\theta}_{H}^{i+1}={\boldsymbol\theta}_{H}^{i}+\alpha\left(\boldsymbol\Phi\left(\boldsymbol\xi_{H}^{i};\mathcal{C}\right)-\boldsymbol\Phi\left(\boldsymbol\xi_{R}^{i};\mathcal{C}\right)\right)$
}
$\mathbf{p}^{*} \leftarrow \textup{Optimize} (\boldsymbol\theta_{HP}^{i+1},\boldsymbol\theta_{RP},\mathcal{C})$

$\mathbf{t}^{*} \leftarrow \textup{Optimize} (\mathbf{p}^{*},\boldsymbol\theta_{HV}^{i+1},\theta_{RV},\mathcal{C})$

$\boldsymbol\xi_{R}=f(\mathbf{p}^{*},\mathbf{t}^{*})$

$\boldsymbol\tau \leftarrow \textup{Impedance}(\boldsymbol\xi_{R})$ 

$i=i+1$
}
\caption{Learning human preferences from kinesthetic demonstration}
\label{Algo1}
\end{algorithm} 

Finally, the trajectory that adheres to both the path and velocity preferences is constructed using $\boldsymbol\xi_{R}=f(\mathbf{p}^{*}_{V},\mathbf{t}^{*})$. The full method is summarized in Algorithm \ref{Algo1}.

%%%%%%%%%%%%%%%%%%%%%%%%%%%%%%%%%%%%%%%%%%
\section{Method Validation with User Study}\label{UserStudy}
\begin{figure}
  \includegraphics[width=\columnwidth]{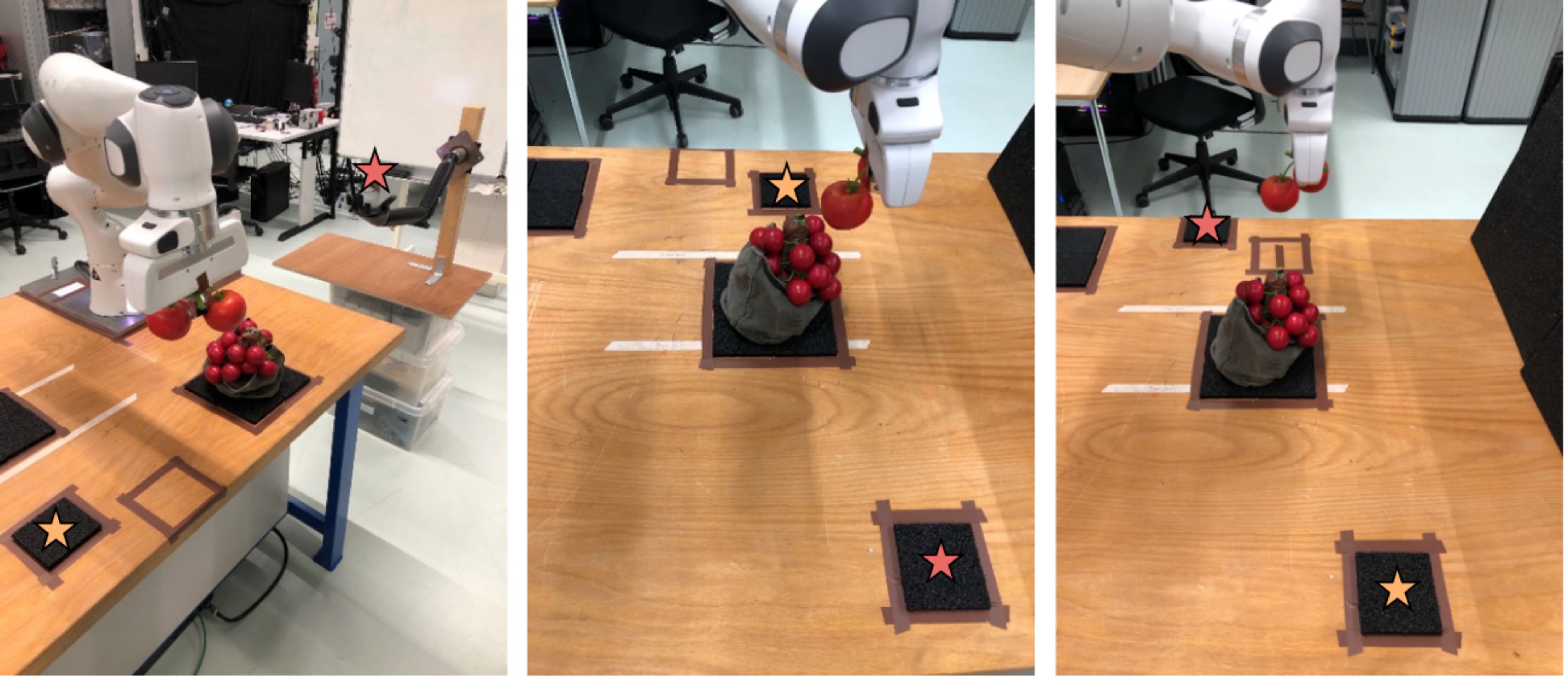}
  \caption{From left to right Scenarios 1--3. The orange and red star, respectively, indicate the start and goal positions. The obstacle to be avoided is the bag of tomatoes. Scenario 1 and 2 shared the same starting positions, and Scenario 2 and 3 shared the same obstacle positions. Notice the difference in height of the goal position in Scenario 1 compared to Scenarios 2 and 3.}
  \label{setup}
\end{figure}

To validate our framework we conduct two user experiments on a Franka Emika 7-DoF robot arm. Thereby we show a proof-of-concept of our approach in a real-world scenario with non-expert users. In both experiments, we use a set of three pick-and-place tasks in an agricultural setting as shown in Fig.~\ref{setup}. The primary goal of each task was moving the tomatoes from the initial position to the goal without any collisions with the obstacle. The experiments were approved by the Human Research Ethics Committee at the Delft University of Technology on 06/09/2021.

We recruited 14 participants (4 women, 10 men) between 23 and 36 years old (mean $= 26.8$, SD $= 3.6$), six of whom had prior experience with robotic manipulators, but none of whom had any exposure to our framework.

\begin{figure*}
    \centering
    \includegraphics[width=0.9\linewidth]{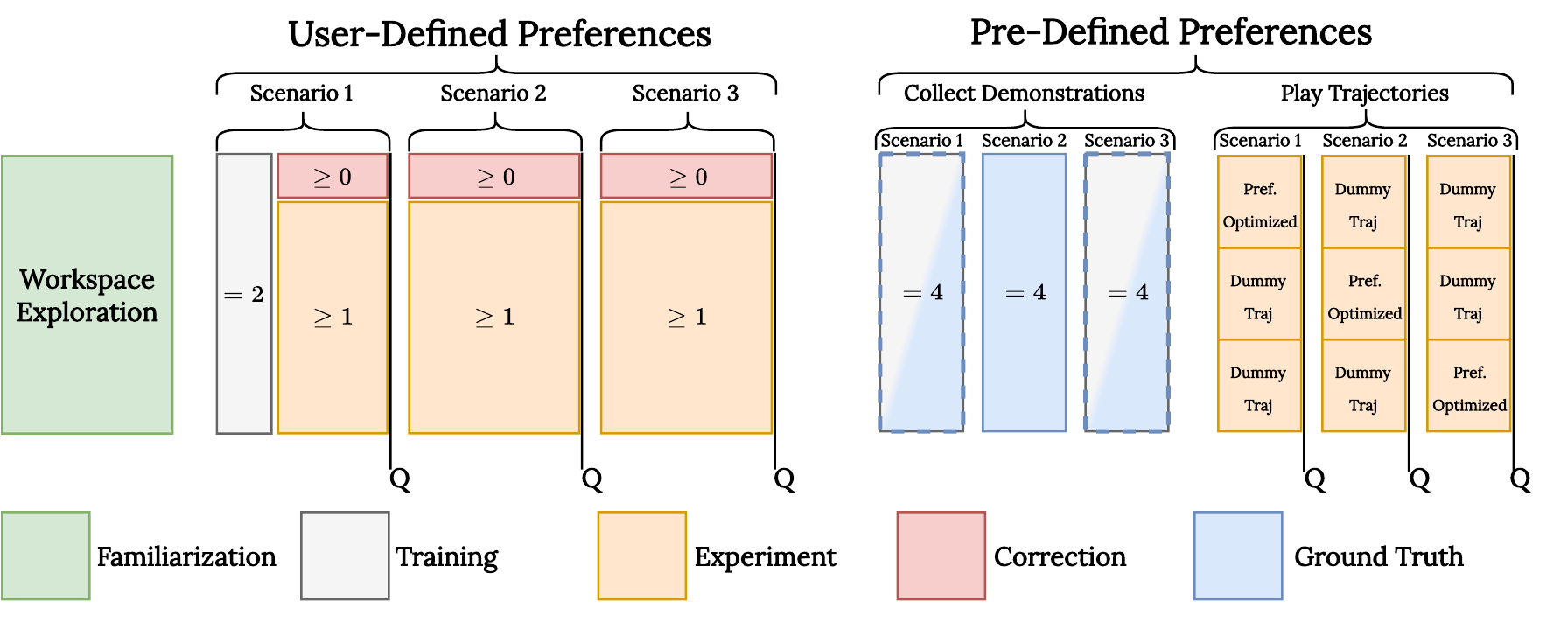}
    \caption{The experimental protocol. Users started with workspace familiarization, then went through the first experiment assessing the performance of the framework in understanding their preferences. Finally, in the last experiment, they provided ground truth demonstrations and evaluated the demonstrated trajectories in adhering to the set of predefined preferences. The numbers indicate the number of demonstrations given, either by the human (training/correction/ground truth) or the robot (experiment). The order in which the dummy trajectories were shown to the users was different in every scenario. The `Q' symbols indicate when participants were provided with questionnaires.}
    \label{protocol}
\end{figure*}

Each user first took approximately 10 minutes to get familiar with physically manipulating the robot in the workspace. In this period, we also instructed users about the goal of the task and the preferences the robot could capture. Users then proceeded with the two experiments. To subjectively assess whether the framework can capture a range of different behaviors, in the first experiment we let the users freely choose their path and velocity preferences. Once users were more familiar with the framework, in the second experiment we assessed how effectively they could teach a set of pre-defined preferences to the robot. The overview of the user study is provided in Fig.~\ref{protocol}. We discuss each experiment in the following subsections. A video of the experiments can be found here: \url{https://youtu.be/hhL5-Lpzj4M}.

\subsection{User-Defined Preferences}\label{user-defined}
In the first experiment, we investigate how our framework performs when users openly choose their set of preferences. We are specifically interested in assessing how well the robot plans motions in new task instances with a context it has not seen before (i.e.\ generalization of preferences to new scenarios). We also evaluate the user experience in terms of acceptability and effort required from the user's perspective. Accordingly, we test the following hypotheses:  

\noindent \textbf{\textit{H1.}} The proposed framework can capture and generalize user preferences to new task instances.

\noindent \textbf{\textit{H2.}} Users feel a low level of interaction effort.

\noindent \textbf{Procedure and Measures.} Users first performed a demonstration in Scenario~1 (Fig.~\ref{setup}) for path preferences with the robot in gravity compensation mode. Notably, we did not limit users to a discrete set of preferences. For instance, instead of asking users to pass on either the close or far side of the obstacle, we asked them to intuitively demonstrate how far to either side of the obstacle they would prefer to pass. They could, for example, decide to pass right above the obstacle which would correspond to a ``stay to the middle of the obstacle'' for the ``obstacle side'' path preference.
We then collected a second separate demonstration for the velocity preferences. During velocity demonstrations, the robot was only compliant along a straight line path covering the full range of distances to the obstacle. This allowed the users to demonstrate their preferred speed without having to care about the path. The velocity optimization step can take up to 3 minutes, therefore we simplified the method for learning and planning velocity preferences to find the velocity $c_j$ with the highest feature count in this part of the study. 
Users were instructed to provide corrections via additional kinesthetic demonstrations (max 10~min per scenario) until they were satisfied with the resulting trajectory. However, the users were informed that the trajectory speed was only trained once and would not be updated further.

\begin{figure*}
    %\centering
    \includegraphics[width=\linewidth]{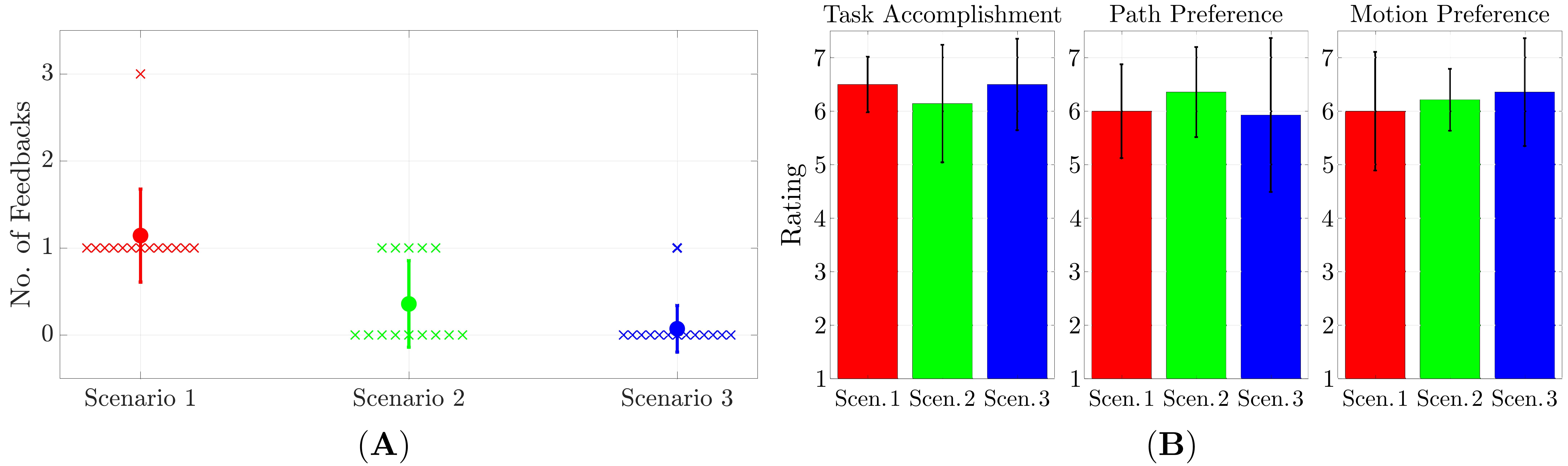}
    \caption{Results of the first experiment. (\textbf{A}) An average number of feedback provided to the system for each task. The dot represents the mean score, the error bars represent the standard deviation, and the crosses indicate individual data points. (\textbf{B}) Results of the Likert questionnaire for the first resulting trajectory in every task (i.e.\ prior to any additional demonstrations) - the error bars correspond to standard deviation.}
    \label{fig:subfigures}
\end{figure*}

After observing each trajectory, the users filled out a subjective questionnaire for qualitative evaluation, rating the following statements on a 7-point Likert scale:
\begin{enumerate}
    \item The robot accomplished the task well. 
    \item The robot understood my \textbf{path} preferences.
    \item The robot understood my \textbf{motion} preferences.
\end{enumerate}
To evaluate the effort, we counted the number of times a user-provided feedback, and let the participants fill out the NASA Task Load Index (TLX) at the end of this experiment. The independent variables of this experiment are the contexts which are varied for each scenario for assessing workload. While we do not compare results with a baseline here, NASA-TLX is still appropriate since it can capture absolute results \cite{hart1988development}. 

\noindent \textbf{Results.} Users demonstrated a multitude of path preferences, including ``Keep low distance to the obstacle'' and ``Stay at medium height above the table''. Similarly, for velocity preferences, while the majority opted for a constant ``medium'' speed, both the preference of going ``slower when close to the obstacle'' and ``faster when close to the obstacle'' were demonstrated at least once.

Fig. \ref{fig:subfigures}(A) shows that the average amount of feedback given to the system after the first task drops, with the majority of the users satisfied with the results of generalization after the initial demonstration (we count the training step in Task~1 as feedback). This result is also reflected in Fig. \ref{fig:subfigures}(B), showing that the users scored the first trajectory produced in every scenario consistently high for all three statements, supporting the claim that the framework can generalize both path and velocity preferences to new task instances. This provides strong evidence in favor of both \textbf{H1} and \textbf{H2}.

\begin{figure}
  %\centering
  \includegraphics[width=\columnwidth]{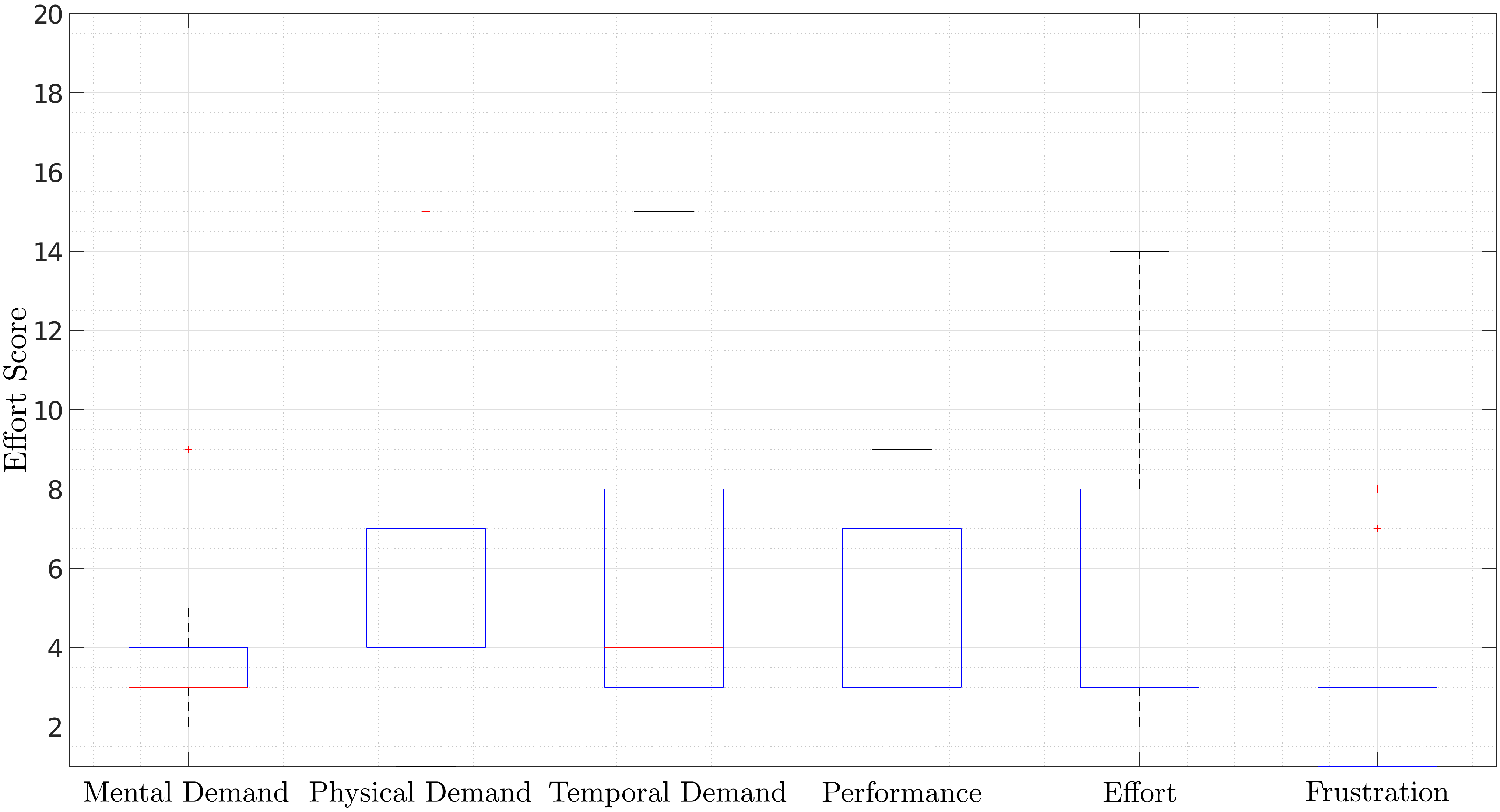}
  \caption{Results of the NASA-TLX questionnaire after the first experiment.}
  \label{TLX}
\end{figure}

The NASA-TLX results in Fig. \ref{TLX} show that the users experienced low mental and physical workload. Although kinesthetic teaching is normally associated with high effort, our framework's effort scores remain mostly on the lower side of the scale. 
One participant was particularly strict on a height preference the algorithm failed to capture, resulting in 3 iterations of feedback in Scenario~1. Overall, the results in Fig.~\ref{TLX} support \textbf{H2}.

\subsection{Pre-Defined Preferences}\label{pre-defined}
To objectively evaluate the accuracy, and the user's ability to discern preferences, we conduct an experiment where users are asked to adhere to the following path preferences (we did not consider velocity preferences in this experiment):
\begin{itemize}
    \item Pass on the side of the obstacle that is closer to the robot.
    \item Stay far from the obstacle.
    \item Keep a high elevation from the table.
\end{itemize} 
Exactly how to express these preferences, and how to trade off between them if necessary, is left to the users.
We test the following hypotheses: 

\noindent \textbf{\textit{H3.}} The method remains consistently accurate in all scenarios.

\noindent \textbf{\textit{H4.}} Users can clearly distinguish that the output of the framework is following the specified preferences.

\noindent \textbf{Procedure and Measures.} We collected four demonstrations per scenario. For half of the participants, we trained the model on the mean of the four demonstrations from the first scenario, and for the other half, we used the mean of data from the third scenario. This is to establish that our method generalizes, even when changing the set used as the training data. 

\begin{figure}
  %\centering
  \includegraphics[width=\columnwidth]{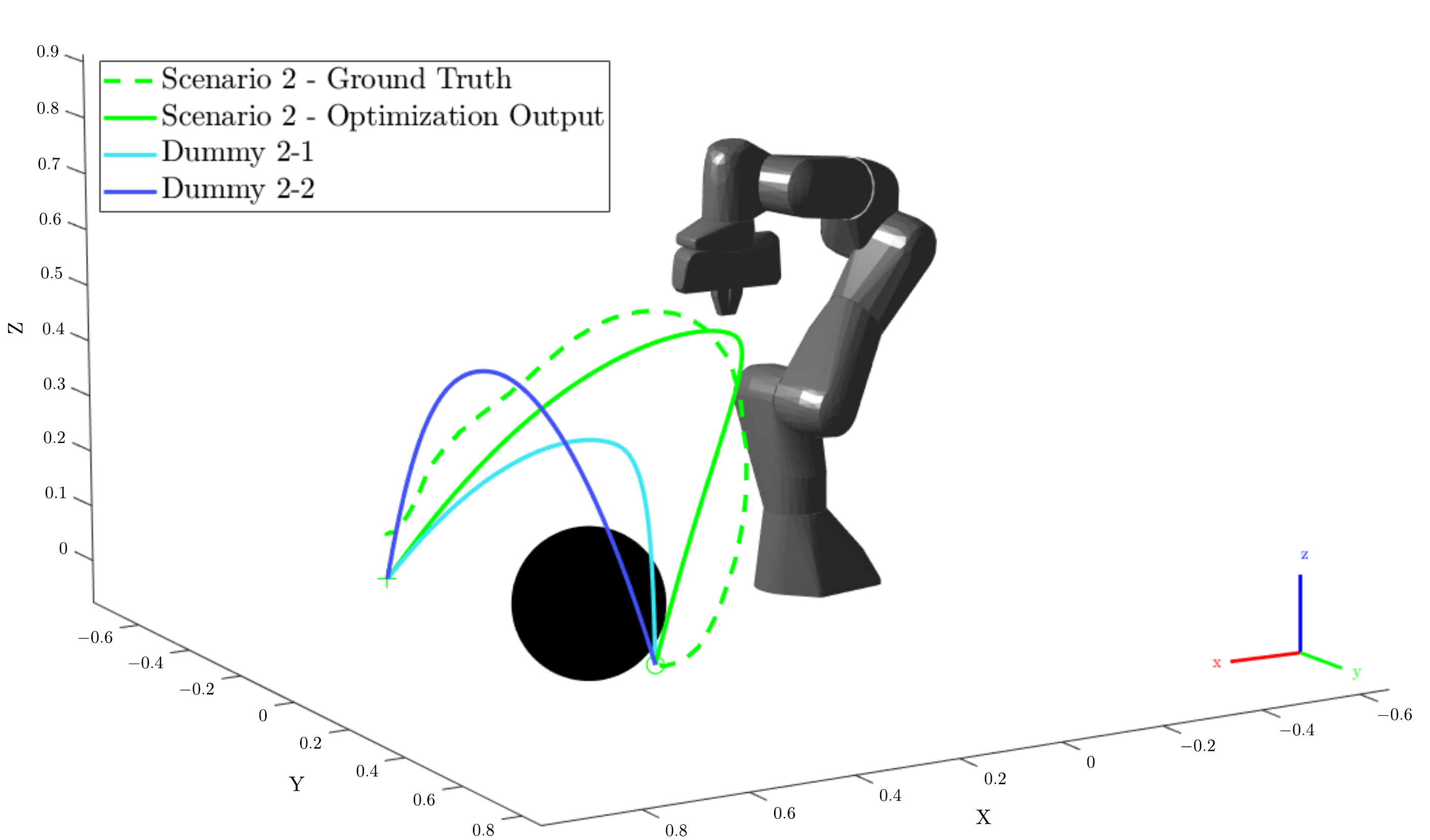}
  \caption{Scenario~2 results (second experiment) for a single user. The dummy trajectories, in light and dark blue, are designed not to meet the `height from table' and `obstacle side' preferences respectively. The green dashed and solid lines are the mean of human ground truth demonstrations and the robot trajectory respectively. The black sphere represents the obstacle. The framework was trained on data from Scenario~3 and had no access to the ground truth shown.}
  \label{trajs}
\end{figure}

After that, the users were shown 3 trajectories per scenario: the output of our framework, and two dummy trajectories (Fig. \ref{trajs}). The dummy trajectories were designed to adhere to 2 out of 3 path preferences. This allowed us to observe if users could distinguish our method’s results compared to sub-optimal trajectories. 

As an objective measure of the accuracy of our method, we computed, per scenario, the total Euclidean distance of samples within each trajectory with respect to the mean of the demonstrations (using $N$=80). Furthermore, we compare the total feature counts along each trajectory and measure the error with respect to the ground truth in the feature space. 

Subjectively, users rated a 7-point Likert scale per trajectory: ``The robot adhered to the demonstrated preferences''.

\begin{table}
\caption{Average distance error of trajectory samples w.r.t.\ the ground truth, normalized w.r.t.\ distance of start to goal, in meters: mean [min, max].}
\label{tabl:distance}
%\centering
\resizebox{\linewidth}{!}{%
\begin{tabular}{ccllcllcll}
\toprule
\multicolumn{1}{l}{} & \multicolumn{3}{c}{Scenario 1}               & \multicolumn{3}{c}{Scenario 2}               & \multicolumn{3}{c}{Scenario 3}               \\ \midrule
Optimized                & \multicolumn{3}{c}{0.14 {[}0.09, 0.18{]}}  & \multicolumn{3}{c}{0.20 {[}0.12, 0.27{]}}  & \multicolumn{3}{c}{0.17 {[}0.13, 0.24{]}} \\
Dummy 1              & \multicolumn{3}{c}{0.24 {[}0.13, 0.34{]}} & \multicolumn{3}{c}{0.26 {[}0.16, 0.38{]}} & \multicolumn{3}{c}{0.30 {[}0.21, 0.41{]}} \\
Dummy 2              & \multicolumn{3}{c}{0.27 {[}0.18, 0.33{]}} & \multicolumn{3}{c}{0.39 {[}0.28, 0.47{]}} & \multicolumn{3}{c}{0.23 {[}0.18, 0.28{]}} \\ \bottomrule
\end{tabular}
}
\end{table}

\begin{figure}
    \centering
    \includegraphics[width=\columnwidth]{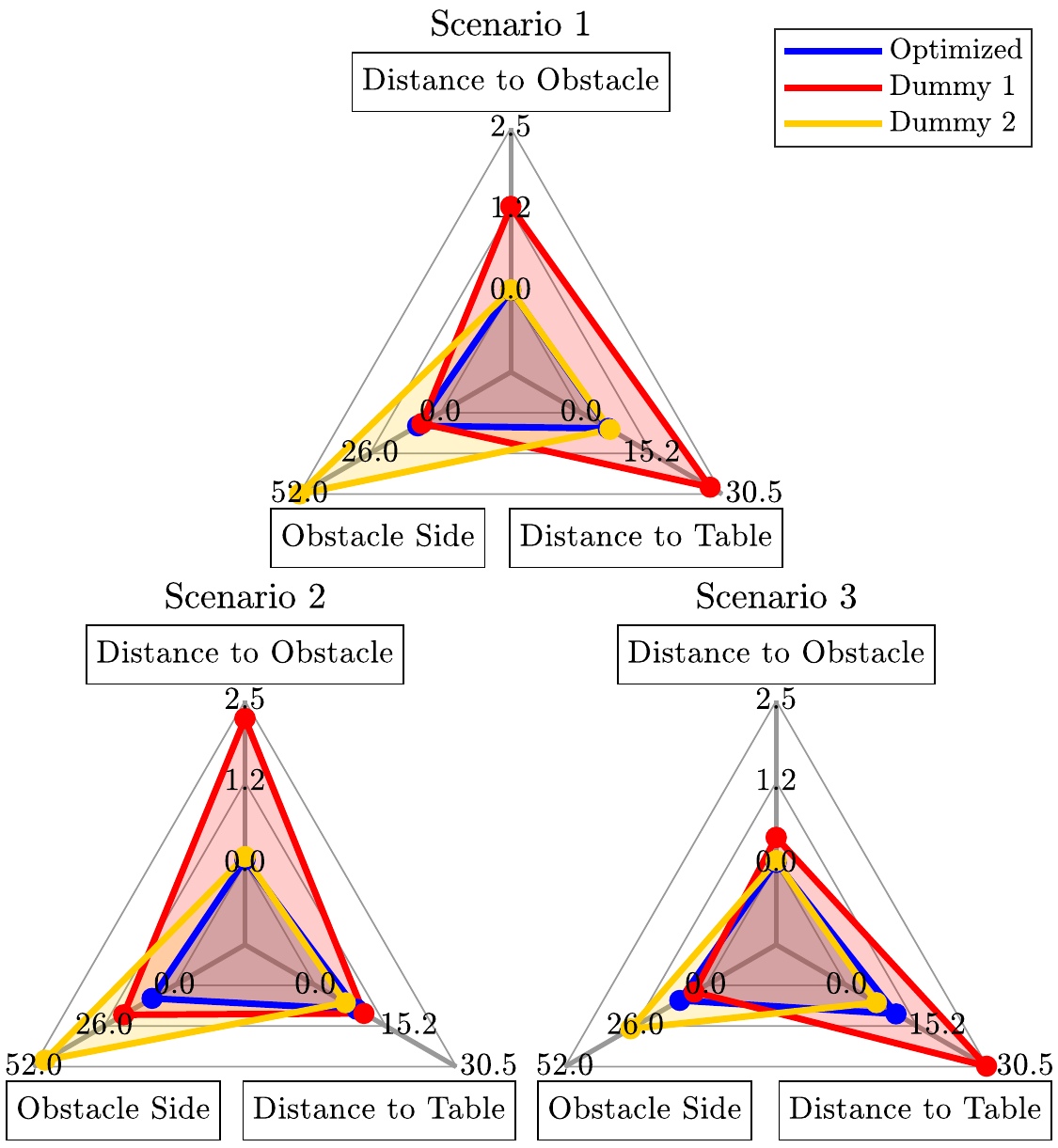}
    \caption{Total feature count errors of each path preference (all participants), w.r.t. the ground truth (i.e.\ smaller values for each axis are favored).}
    \label{spider}
\end{figure}

\begin{figure}
    %\centering
    \includegraphics[width=\columnwidth]{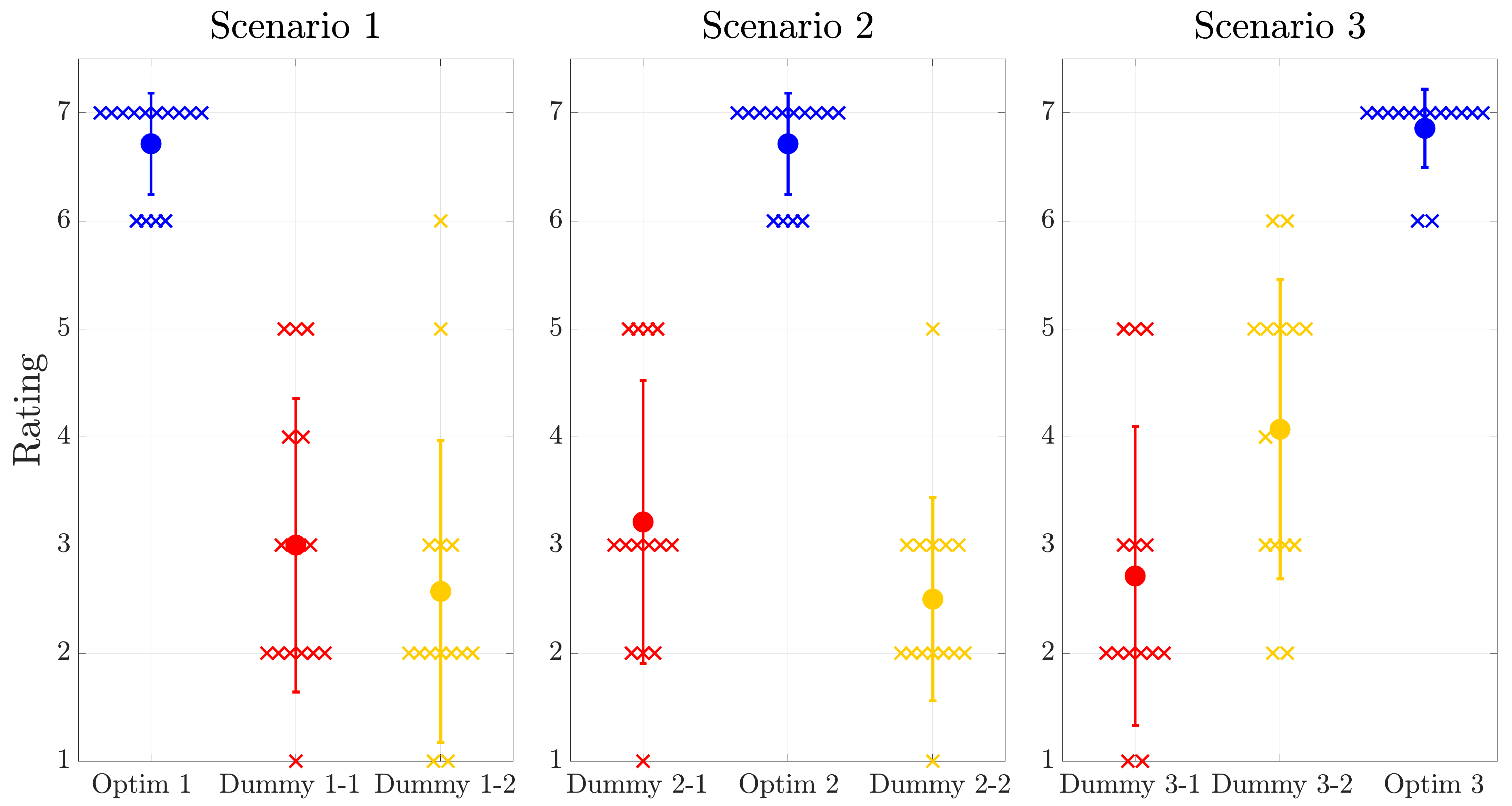}
    \caption{Result of Likert questionnaire for experiment 2. Crosses indicate individual ratings, while the dot and error bar, respectively, represent mean and standard deviation. Users clearly recognize and highly rate the output of the framework in terms of adhering to path preferences.}
\label{Likert2}
\end{figure}

\noindent \textbf{Results.} Fig.~\ref{trajs} shows a generalization result of our method under the aforementioned path preferences. The robot attempts to capture and optimize for each user's personal interpretation of the preferences (e.g.\ one user's definition of `high' is different from another). 
We show the combined results of all users in Tab.~\ref{tabl:distance}, listing the trajectories' mean, min and max Euclidean distance to the ground truth, normalized relative to the start-to-goal distance in each scenario (respectively 1.08, 0.74, 0.88 m). The optimized trajectories have the smallest error, but the results only partially support \textbf{H3}, as the error in Scenario 2 and 3 is slightly larger than in Scenario 1. This scenario has the longest distance from start to goal, for which the framework seems to perform better.

Fig.~\ref{spider} shows the errors of the trajectories in feature space. In all scenarios, our optimization result occupies the smallest area. However, in Scenario 2 and 3 the optimized trajectories occupy a slightly larger area than in Scenario 1, showing the same trend of performance loss in scenarios with the shorter length. Furthermore, in Scenario 2 and 3, dummy trajectories occasionally perform slightly better for one of the preferences. Nevertheless, we see in Fig. \ref{Likert2} that users clearly score the output of our framework higher, which strongly supports \textbf{H4}. This indicates that users prefer all preferences to be satisfied simultaneously. 
The best performing dummy (S3-D2), with the smallest area in Fig.~\ref{spider} and lowest values in Tab.~\ref{tabl:distance}, correlates to a high rating in Fig.~\ref{Likert2}. This also supports \textbf{H4}, suggesting that non-expert users intuitively recognize such preferences in trajectories.

\subsection{Discussion} \label{discussionUS}
As the state-of-the-art methods do not have the same functionalities (e.g., path-velocity separation) as the proposed method, we conducted a user study only on the proposed method itself. To account for that, we employed absolute types of metrics (i.e., Likert and NASA-TLX), which can be interpreted independently, rather than tied to a specific external baseline. For example, the Likert scale is tied to an agreement with the given statements and the natural point on the agreement scale serves as a general baseline. The advantage of this is that the result is not tied to a specific relative baseline. If methods that enable the same functionalities are developed in the future, the same Likert scale/questionnaire can be employed to compare the subjective results independently of a specific baseline.

An advantage of the proposed method is that it learns fast. During the first part of the user study, participants spent on average 16.5~s interacting with the robot before expressing satisfaction with the results. This is partially due to having access to kinesthetic demonstrations. This method of demonstration has been criticized as challenging in applications involving high DoF manipulators \cite{jain2015learning,akgun2012keyframe}. However, the separation of learning and control in our framework means that users do not have to provide the correct configuration of the arm in their demonstrations. This feature made it significantly easier for the users to provide demonstrations, which is reflected in the reported low mental and physical loads (Fig.~\ref{TLX}).

The separation of path and velocity planning has additional benefits. Formulating the optimization as a multi-objective problem with both position and velocity features results in undesirable interaction of objectives. For instance, when velocity features reward high speeds, the trajectory would converge to a longer path. Conversely, path features with high rewards in specific regions of space would result in slow motion in those regions to increase the density of samples and consequently the overall reward. On the other hand, the separated trajectory optimization has the limitation that it cannot account for dynamical quantities such as joint velocity and acceleration, and the efficiency of movements in robot's joint space can not be considered.

A challenge with our definition of robot and user objectives is that the trajectory optimization outcome does not always align with task requirements. For instance, a strong ``stay close to the object'' preference can result in a minimum cost for a trajectory that is briefly in a collision. Tuning the collision weight can only partially solve this issue, as at a certain point this cost can interfere with the path preferences.

Our user study results showed that non-expert users can intuitively use our method to quickly teach a wide range of preferences to the robot. While the generalization results to different task instances show that we do not always reproduce trajectories with the exact desired shapes in the workspace (see Fig.~\ref{trajs}), the subjective performance evaluation shows that users still deem these trajectories highly suitable in terms of task accomplishment and the preferences achieved. State-of-art LfD methods are very capable of producing accurate and complex dynamic movements \cite{mulling2013learning}. However, in tasks where there are multiple ways of achieving the same goal, we prefer to trade off motion accuracy for achieving planning propensities on a higher level.

Unfortunately, our approach inherits the limitations of IRL approaches that require specifying reward features by hand. Both features and robot rewards depend on several parameters which require tuning. The problem becomes especially difficult as our features simultaneously govern the behavior of reward learning and trajectory optimization. For instance, high gradients in the feature function lead to erratic behavior of the optimizer, leading to poor solutions and convergence to local optima. Yet, for certain features, a sufficiently high gradient is required to facilitate the learning of preference weights that are large enough to counterbalance each other. As a result, we had to resort to further tuning of parameters, such as the learning rate in \eqref{Update}. An interesting direction for future work would be to test if and how well these issues can be alleviated by feature learning from additional demonstrations, as was done in \citet{bobu2022inducing}. Furthermore, an approach in \cite{katz2021preference} could be employed to learn the relative weighting among features and add additional features through nonlinear functions using neural networks.

In feature engineering or learning, the definition of the context determines how expressive the features are. We considered a limited set of vectors as the context in this work (i.e.\ obstacle position, start and goal positions). It is possible to include additional information, such as object properties (e.g.\ sharp, fragile or liquid) \cite{jain2015learning}, human position \cite{bajcsy2017learning,losey2022physical}, and number of objects. The more rich the context, the more preferences the model can capture in complex environments. However, training diversity can become an issue with contextually rich features, as the model would require more demonstrations to cover a wider range of situations. This will increase training time. An evaluation of the trade-off between improved generalization and higher training time is left for future work.

%%%%%%%%%%%%%%%%%%%%%%%%%%%%%%%%%%%%%%%%%%
\section{Supplementary Comparison Study} \label{ComparisonStudy}
The purpose of this supplementary study is to highlight different aspects and properties of our method in comparison to two common methods from the literature: Dynamic Movement Primitives (DMPs) \cite{Calinon19chapter} modified with potential fields for obstacle avoidance \cite{gams2016adaptation}, and the method used in \citet{bajcsy2017learning} (referred to as PHI). Since these methods are different conceptually and by design (i.e., optimize for different properties), quantitative comparison is not meaningful. Thus, we examine their aspects in a practical transportation task qualitatively. These aspects are: adherence to preferences, robot objectives, trajectory feasibility, and online learning. In the next subsections, we first discuss the different aspects in more detail, before showing the effects in the transportation task and discussing the pros and cons of the different methods.

\subsection{Conceptual differences per aspect} \label{differences}

\subsubsection{Adherence to preferences}
The methods capture preferences in a different way. Even though we added obstacle awareness to the DMPs we compare to, they lack an explicit notion of preferences. A forcing function is learned to match the shape and velocity distribution of the demonstration, but without any parameterization over features that may capture behavior relative to the context. The potential fields for obstacle avoidance add a basic level of context-awareness, but a predefined one.

Both our method and PHI learn an explicit preference model that is structured as a linear combination of context-parameterized features. Like our model, PHI considers the ``\emph{Height from the Table}'' and ``\emph{Distance to the Obstacle}''. We additionally consider the ``\emph{Obstacle Side}'', such that our features cover the different dimensions in space and allow us to capture the preferences in every direction. PHI instead considers other features, such as ``\emph{Distance to Human}'' and ``\emph{Efficiency}''. 

Our features are counterbalanced by explicit robot objectives (Sec.~\ref{sssec:robot-objectives}). In PHI, it is possible to replace the features with the features we use, including the ones for the robot which will not be updated during learning. This way, we can test the effects of the change of features and the change of method.

\subsubsection{Robot objectives} \label{sssec:robot-objectives}
In contrast to PHI, we chose to explicitly separate objectives such as ``\emph{Path Efficiency}'' and ``\emph{Collision Avoidance}'', from the preferences we try to capture. Instead, we let the robot have a reward function of its own. The same effect can be achieved in PHI by fixing the weights of selected features. 

The effects of the trade-off between the learned human rewards and the given robot objectives visible in the iterative updates can be viewed as a negotiation between the preferences of two independent agents. We believe that this separation and negotiation will be beneficial especially as tasks become more complex and the artificial agent has knowledge complementary to the human. The benefits may be less visible in the simple task considered in this paper.

As DMPs do not explicitly model an objective function to be optimized, this attribute does not apply.

\subsubsection{Trajectory feasibility}
Our method does not automatically check if the planned trajectory is feasible to execute by the robot. A motion feasibility objective can be added to the robot objective function to take this into account in the path optimization.

Rather than weighing the learned preferences against robot objectives, PHI ensures motion feasibility by optimizing the trajectory in the robot configuration space. This requires an additional simulation step, incorporating the kinematic model of the robot, during trajectory optimization. But then, no corrections need to be applied in hindsight to ensure trajectory feasibility.

\subsubsection{Online Learning}
Our method requires a full trajectory to learn from, whereas PHI updates the internal model at each time step. This potentially makes our method less efficient. On the other hand, it allows us to capture velocity preferences in addition to path preferences. Also, because we separate the demonstration from the execution, we obtain a more `clean' observation of the preferences, as we do not have to deduce from the interaction forces what the human demonstration would have looked like without the robot interference. This is likely to benefit generalization. In case of large user corrections, it may even reduce the user's effort to demonstrate their preferences without the robot interfering. Especially velocity preferences have been found cumbersome to correct in an online manner \citep{losey2019learning}.

When it comes to online updates, DMPs are the fastest, because there is no complex underlying model that needs to be updated. But the trade-off of having a much simpler model is that it lacks the ability to capture preferences in a way that might generalize to changes in the scenario.

All three methods update their model to reduce the error with respect to the latest observation from demonstration. A learning rate trades off learning and overfitting on the corrective demonstration. DMPs updates correct the behavior on the trajectory level, whereas PHI and our method update at a higher level where the observed trajectories are considered a consequence of a human reward model. Nevertheless, future observations that appear contradictory to earlier ones will cause (partial) unlearning of the earlier updates. This results in erroneous behavior learned from imperfect corrections to be corrected, but may in some cases also lead to undesired unlearning.

\subsection{Comparison}
We will now present a qualitative comparison between the three methods, PHI with two different feature sets: \_$\phi_\text{orig}$, \_$\phi_\text{our}$. Our aim is to show the effects of the conceptual differences discussed in Sec.~\ref{differences}. To make the comparison as fair as possible, we let all the models learn from the same demonstration data. All methods have access to and consider the obstacle position for planning.

\begin{figure}[t]
  %\centering
  \includegraphics[width=\columnwidth]{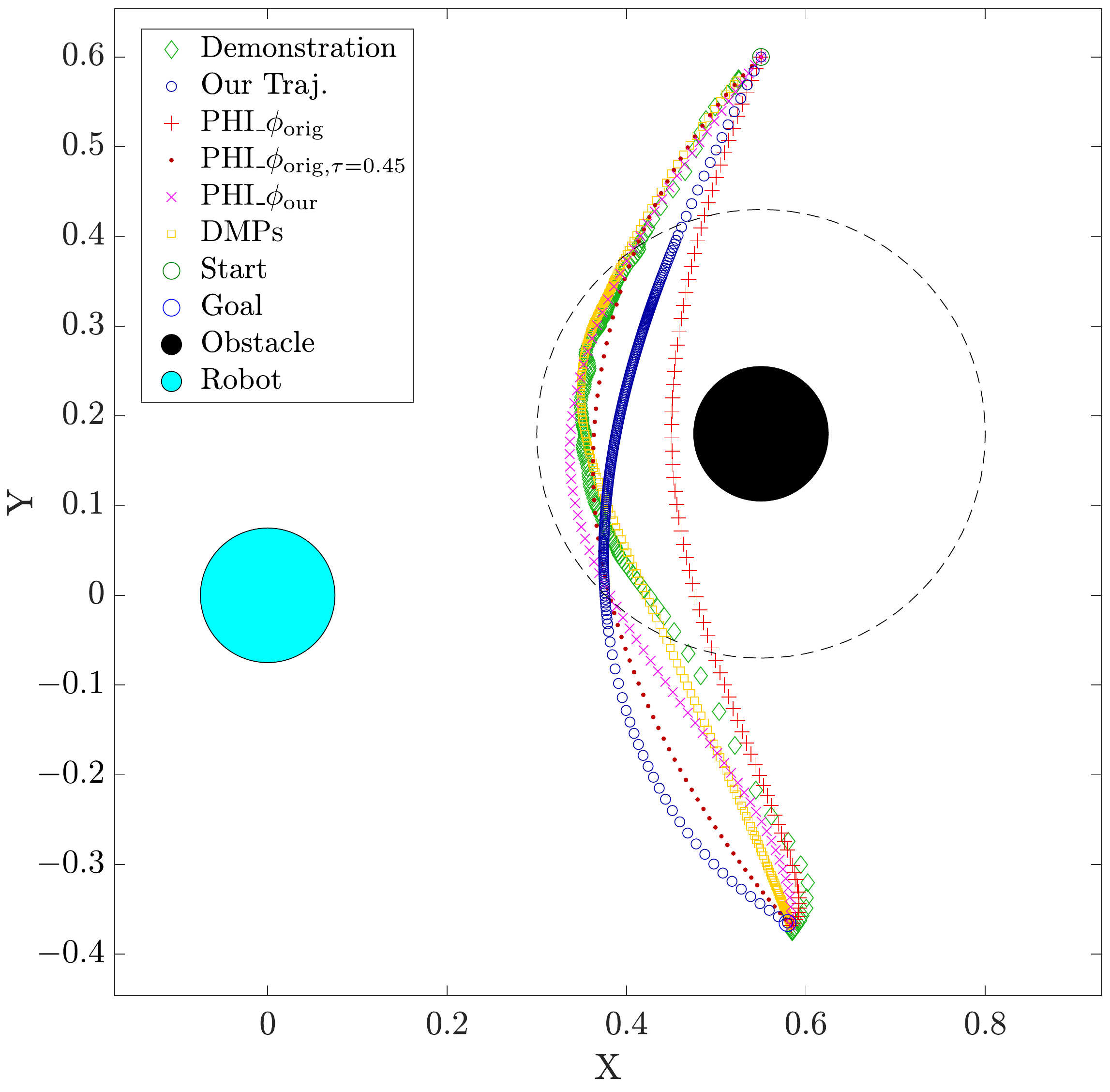}
  \caption{Training scenario with the human demonstrated trajectory (green diamonds) and the learned reproductions: ours in dark blue circles, PHI\_$\phi_\text{orig}$ in red plus signs, with an intermediate learning result in dots, PHI\_$\phi_\text{ours}$ in purple crosses, and DMP in yellow squares. By placing the markers at equal time intervals, we display the velocity on the trajectories (i.e.\ the closer the markers, the slower the motion). As PHI does not support differences in velocity, all red and purple markers are spaced equally along the trajectory. The black, cyan, blue, and green circles respectively represent the obstacle, robot, goal (bottom), and start (top) positions. For this study, we set $d_c$= 22.5~cm (indicated by the dashed circle). We consider points within this region as ``close'' to the obstacle.}
  \label{initialScenario}
\end{figure}

We modify PHI to bypass the estimation of the human desired trajectory from forces, as we have direct access to the desired trajectory from demonstration. We compute the ``human correction'' every time step from the mismatch between the planned trajectory and the demonstration. 
The trajectory optimization in PHI requires a robot model for the optimization. As our trajectory optimization does not take the robot dynamics into account, we use a fully actuated point mass for the trajectory optimization. In order to achieve comparable smooth optimal paths, we interpolated the trajectory with a spline instead of linearly as was done originally. For both sets of features, the feature weight ranges and update rates were hand-tuned to achieve as close a trajectory match in the initial scenario as we could manage. This initial scenario is illustrated in Fig.~\ref{initialScenario}.

We consider a situation where the user has a preference for ``passing on the close side of the obstacle'' due to the existence of a wall on the other side that the robot is not aware of. Furthermore, we want to ``remain close to the obstacle'', and to ``slow down when passing close to the obstacle''. We use a single kinesthetic demonstration containing these three preferences as the input to all methods. For PHI\_$\phi_\text{orig}$, we obtained the correct choice of obstacle side in Fig.~\ref{initialScenario} by assuming a person standing on the other side of the obstacle and making use of their ``human feature'', learning not to come too close to the human.

Fig.~\ref{initialScenario} shows the demonstration we use for training, as well as the trajectories obtained from the three methods. As the results are generated for the same context as in the demonstration, these results reflect the performance prior to any generalization of preferences. As PHI updates its internal model at every time step, we observe partial unlearning of some features towards the end of the trajectory. This is particularly visible for the ``Obstacle Distance'' in PHI\_$\phi_\text{orig}$. In Fig.~\ref{initialScenario}, we show an additional trajectory PHI\_$\phi_{\text{orig},\tau=0.45}$, which is generated by PHI with the original features and the weights learned at 45\% of the trajectory. We see that PHI\_$\phi_{\text{orig},\tau=0.45}$ is considerably closer to the demonstrated trajectory. The demonstrated trajectory has many waypoints close together, quite close to the obstacle, as it slows down when passing it. PHI, on the other hand, has its waypoints equally spaced. As a result, towards the end of the trajectory, a considerable batch of PHI waypoints is further away from the obstacle by default. When the weights continue to update on the difference, we obtain the trajectory PHI\_$\phi_\text{orig}$, which lies closer to the obstacle. With our features, in PHI\_$\phi_\text{our}$, the effect is less pronounced as the features trade off differently, yet the learned path is still different from Our Trajectory, as PHI uses a different trajectory optimization method.

Especially considering PHI\_$\phi_\text{our}$, all three methods perform reasonably well in terms of adhering to the aforementioned path preferences, with a slight variation in how close the robot passes by the obstacle.
As discussed in Sec.~\ref{differences}, PHI is not able to capture any velocity preferences. Notably, DMP performs well in this aspect as it is able to replicate the demonstrated behavior in terms of both path and velocity.

\begin{figure}
  %\centering
  \includegraphics[width=\columnwidth]{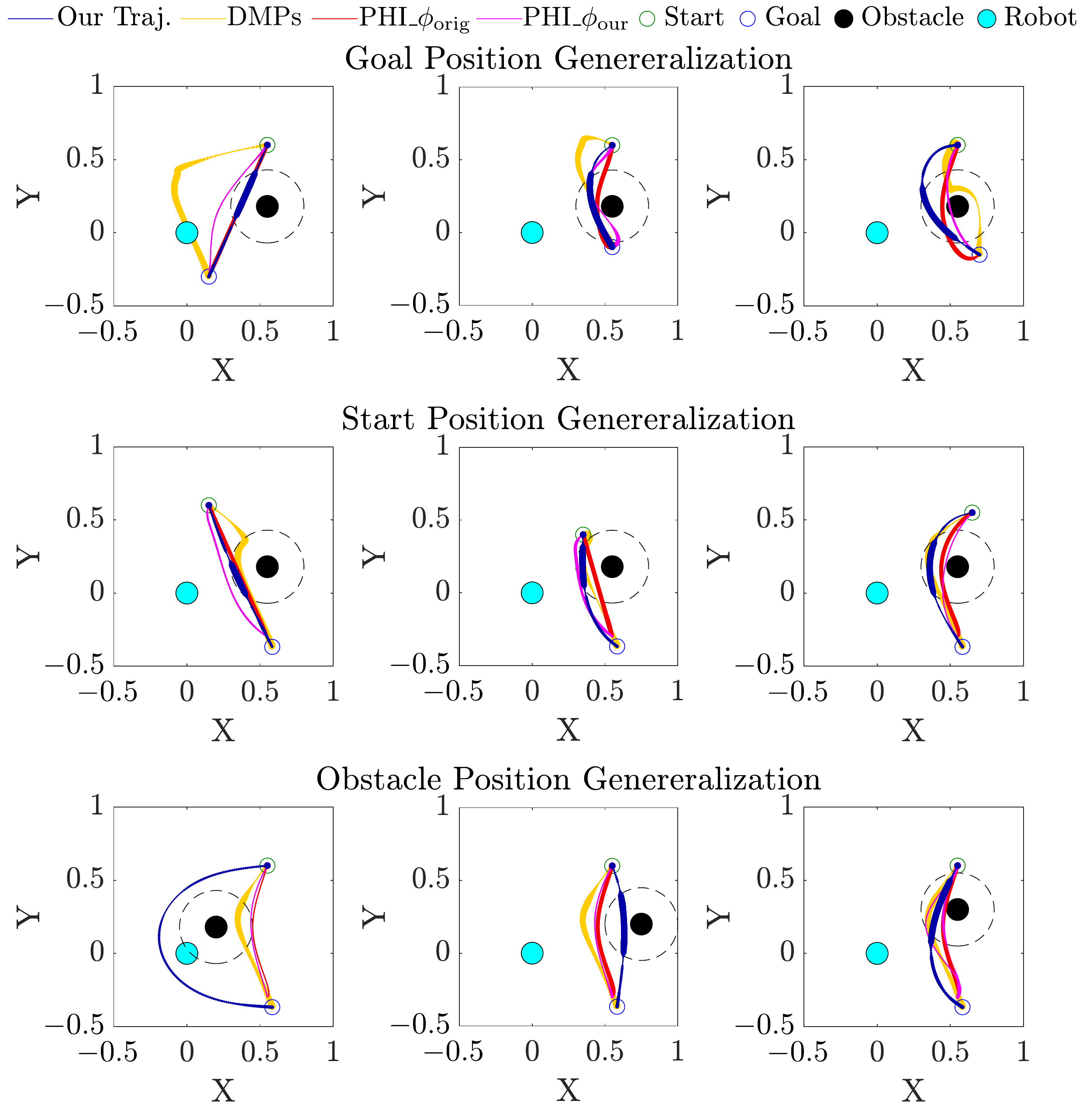}
  \caption{We demonstrate generalization by modifying the goal (top), start (middle), and obstacle (bottom) positions. The yellow, blue, and red and purple trajectories correspond respectively to the output of the DMPs, our framework, and the two versions of PHI. The thickness of the line indicates the inverse of normalized velocity (i.e.\ the thicker the line, the slower the trajectory).}
  \label{DMP}
\end{figure}

\begin{table*}
\caption{Qualitative evaluation of the different aspects of the three methods: DMPs, PHI, and ours. The marker `o' indicates a value between `-' and `+'.}
\label{tab:compare}
\begin{center}
\resizebox{0.8\textwidth}{!}{%
\begin{tabular}{l cccc}
\toprule
 & Adherence to preferences$^a$ & Robot objectives$^b$ & Trajectory feasibility$^c$ & Online learning$^d$ \\ \midrule
DMPs & - & - & - & + \\
PHI  & o & o & + & o \\
Ours & + & + & o & - \\ \bottomrule
\end{tabular}
}
\end{center}

\footnotesize{$^a$The criteria, based on Fig.~\ref{DMP}, where `-' is given for adherence to only a few, or inconsistently many, preferences, and `+' for adherence to most preferences in most of the cases.}

\footnotesize{$^b$The criteria, based on the model structure, where `-' is given when no robot objectives can be added, and `+' when arbitrary robot objectives can be added.}

\footnotesize{$^c$The criteria where `-' indicates no guarantees for trajectory feasibility, and `+' indicates trajectory feasibility can be guaranteed at all times.}

\footnotesize{$^d$The criteria where `-' indicates the inability to learn in real-time, and `+' indicates the ability to learn and re-plan while the task is being executed.}
\end{table*}

Next, we modify the scenario nine times, in three different ways: changing respectively the goal, start, and obstacle positions. We compare how each method is able to generalize the initial observation to the different contexts.
Fig.~\ref{DMP} displays trajectories produced by the three methods in the nine new scenarios, PHI with the two different feature sets.

We observe that the trajectory by our method (shown in dark blue) passes on the left side of the obstacle, close to the robot, in every case; and the velocity preference, of slowing down when passing close to the obstacle, is only achieved by our framework.

\subsection{Discussion} \label{discussionCS}
The comparison with DMPs illustrates how a lack of higher-level knowledge about why a trajectory was demonstrated in a specific manner leads to failure in generalization to new contexts. These results emphasize the need for consideration of human models, such as our reward in \eqref{reward}, in LfD methods. 
PHI, with its model, does considerably better. However, we observe that the internal trajectory optimization reacts differently to the different sets of features, resulting in slight differences in generalized trajectories. The main point, regardless of the applied features, remains that PHI is not able to capture velocity preferences. 
Tab.~\ref{tab:compare} summarizes the strengths and weaknesses of the three methods with respect to the aforementioned aspects. 

PHI optimizes the trajectory in the joint space, which can be done fast since inverse kinematics is only required at waypoints. It ensures the planned trajectories are feasible for the robot, which can be interpreted as implicit robot objectives being satisfied. On the other hand, our method optimizes the trajectory in task space, thus additional inverse kinematics computations are necessary together with an explicit description of corresponding robot objectives. The use of inverse kinematics can also be problematic when there are redundant DoF or when there are potential self-collisions. Nevertheless, planning in the task space is closer to where the human preferences typically are (i.e., more intuitive) and can handle obstacle avoidance in a manner that is more predictable for a non-expert human. 

It should be noted that our framework does take up to two minutes of optimization (total for path and velocity), whereas the DMPs' trajectory is produced instantly. However, there is no guarantee that the DMPs will encode and generalize the desired preferences.

%%%%%%%%%%%%%%%%%%%%%%%%%%%%%%%%%%%%%%%%%%
\section{Conclusion and Future Work} \label{Conclusion}
We presented a novel approach for learning and executing human preferences in robot object-carrying tasks. Our user study showed fast convergence of the algorithm, and a proof-of-concept for generalizing path and velocity preferences. The efficiency and accuracy of our approach were validated in a real-world scenario. Our supplementary study compares the performance of our framework to two common methods from the literature, providing additional insights into the benefits and drawbacks caused by the structural differences between the methods. Both in the user study and in the supplementary study, a single informative feedback sufficed (in all cases except one) to capture the human preferences. In the user study, this was tested without prescribing a preference to the users. Our framework was in most cases successful in generalizing these preferences to previously unseen scenarios. Our results support that our model contributes to personalized planning of object-carrying tasks with low interaction effort.

Future studies comparing our method (with just path preferences) to PHI \citep{bajcsy2017learning,losey2022physical} in a user study could lead to useful insights into people's preferences on iterative versus online learning. Further research could consider a combination of our method and PHI that would benefit from the advantages of both, namely achieving generalization both in-task and over new task instances through learning from online interaction.
Next to that, the trajectory model we used to make the problem tractable is quite simplistic and does not describe human motion behavior very well. Future research can aim to replace this model with a library of motion primitives generated from demonstrations to better capture the shape of the trajectories. More accurate trajectory models can enable the extension of the framework to settings where the human and robot come into contact with each other through a shared object (physical human-robot collaboration). Furthermore, it should be studied whether more complex non-linear formulations of the reward function using Gaussian Processes \cite{biyik2020active} or Neural Networks \cite{ibarz2018reward}, and/or learning them from user input \citep{bobu2022inducing,katz2021preference}, can effectively capture context-aware preferences without the need for rigorous feature engineering.
We believe the presented framework is especially effective in collaborative settings where knowledge of the preferences of a partner is essential to the execution of the task.

\section*{Author Contributions}
A.A., L.v.d.S., L.P., and J.K. developed the concept and methods. The programming was mainly done by A.A., while L.v.d.S. modified PHI for the supplementary comparison study. A.A. and L.v.d.S. contributed to the experiments and data analysis and wrote the first draft of the paper. L.v.d.S., L.P., and J.K. revised the paper. All authors read and approved the submitted version.

\section*{Funding}
The work was partially supported by the European Research Council Starting Grant TERI “Teaching Robots Interactively” (project reference 804907) and European Space Agency through the project “Rhizome: Off-Earth Manufacturing and Construction”. This study received funding from the Honda Research Institute Europe.

\section*{Institutional Review Board Statement}
The study was conducted in accordance with the Declaration of Helsinki, and approved by the Human Research Ethics Committee at the Delft University of Technology on 06/09/2021.

\section*{Informed Consent Statement}
Informed consent was obtained from all subjects involved in the study.

\section*{Data Availability Statement}
The datasets generated for the performed user and supplementary study are available on request. Please contact the corresponding author. 

\section*{Acknowledgments}
The authors thank Andreea Bobu for her feedback on the paper.

\section*{Conflicts of Interest:}
The authors declare no conflict of interest. The funders had no role in the design of the study; in the collection, analyses, or interpretation of data; in the writing of the manuscript; or in the decision to publish the results.

\addtolength{\textheight}{-15.2cm}

\bibliographystyle{apalike} 
\bibliography{references}

\end{document}